\documentclass[conference]{IEEEtran}
\IEEEoverridecommandlockouts
\usepackage{cite}
\usepackage{amsmath,amssymb,amsfonts}
\usepackage{algorithm}
\usepackage{algorithmic}
\usepackage{subcaption}

\usepackage{graphicx} 
\usepackage{booktabs} 
\usepackage{textcomp}
\usepackage{xcolor}
\usepackage{hyperref}
\usepackage{xcolor,tcolorbox}
\usepackage{amsmath}

\def\BibTeX{{\rm B\kern-.05em{\sc i\kern-.025em b}\kern-.08em
    T\kern-.1667em\lower.7ex\hbox{E}\kern-.125emX}}
\begin{document}

\title{English Please: Evaluating Machine Translation with Large Language Models for Multilingual Bug Reports}

\author{\IEEEauthorblockN{1\textsuperscript{st} Avinash Patil}
\IEEEauthorblockA{
\textit{Juniper Networks Inc.}\\
Sunnyvale, USA \\
patila@juniper.net \\ ORCID: 0009-0002-6004-370X}
\and

\IEEEauthorblockN{2\textsuperscript{nd} Siru Tao}
\IEEEauthorblockA{
\textit{Juniper Networks Inc.}\\
Sunnyvale, USA \\
stao@juniper.net \\
ORCID: 0009-0001-2243-7807}

\and
\IEEEauthorblockN{3\textsuperscript{rd} Aryan Jadon}
\IEEEauthorblockA{
\textit{Juniper Networks Inc.}\\
Sunnyvale, USA \\
aryanj@juniper.net \\
ORCID: 0000-0002-2991-9913}
}


\maketitle

\begin{abstract}
Accurate translation of bug reports is critical for efficient collaboration in global software development. In this study, we conduct the first comprehensive evaluation of machine translation (MT) performance on bug reports, analyzing the capabilities of DeepL, AWS Translate, and large language models such as ChatGPT, Claude, Gemini, LLaMA, and Mistral using data from the Visual Studio Code GitHub repository, specifically focusing on reports labeled with the \texttt{english-please} tag. To assess both translation quality and source language identification accuracy, we employ a range of MT evaluation metrics—including BLEU, BERTScore, COMET, METEOR, and ROUGE—alongside classification metrics such as accuracy, precision, recall, and F1-score. Our findings reveal that while ChatGPT (gpt-4o) excels in semantic and lexical translation quality, it does not lead in source language identification. Claude and Mistral achieve the highest F1-scores (0.7182 and 0.7142, respectively), and Gemini records the best precision (0.7414). AWS Translate shows the highest accuracy (0.4717) in identifying source languages. These results highlight that no single system dominates across all tasks, reinforcing the importance of task-specific evaluations. This study underscores the need for domain adaptation when translating technical content and provides actionable insights for integrating MT into bug-triaging workflows. The code and dataset for this paper are available at GitHub—\url{https://github.com/av9ash/English-Please}.

\end{abstract}


\begin{IEEEkeywords}
Multilingual Bug Reports, Machine Translation (MT), ChatGPT, Claude, Gemini, LLaMA, Mistral, AWS Translate, DeepL, Visual Studio Code, \texttt{english-please}, BLEU, BERTScore, COMET, METEOR, ROUGE, Software Development, Domain-Specific Translation, Technical Text Localization, Large Language Models, Source Language Identification
\end{IEEEkeywords}

\section{Introduction}

As global software development increasingly depends on multilingual collaboration, the accurate translation of technical documents has become a crucial challenge. Among these, bug reports present unique difficulties due to domain-specific jargon, software version references, and embedded code snippets—elements that often challenge generic machine translation (MT) systems~\cite{chu2018survey}. While recent advances in neural MT have significantly improved translation fluency and accuracy~\cite{bahdanau2014neural, Vaswani2017}, their effectiveness on specialized technical texts, such as software bug reports, remains an open question.

This study examines bug reports from the official Visual Studio Code (VS Code) GitHub repository~\cite{VSCodeRepo}, specifically those labeled \textit{english-please}. This tag indicates that the original report was submitted in a language other than English and requires translation for broader accessibility. Ensuring accurate translations of these reports is essential—minor errors or misinterpretations can obscure the nature and severity of a bug, hindering efficient debugging and software maintenance~\cite{arnaoudova2016linguistic}.

We evaluate the translation and language identification performance of several leading MT systems: ChatGPT, AWS Translate, DeepL, Claude, Gemini, Mistral, and LLaMA. Each system is tasked with identifying the source language and translating non-English bug reports into English. Language identification accuracy is especially critical, as an incorrect assumption about the source language can cascade into flawed translations. Our evaluation of language detection is based on four established metrics: accuracy, precision, recall, and F1-score.

To evaluate translation quality, we use five complementary metrics: \textit{BLEU}\cite{papineni2002bleu}, \textit{BERTScore}\cite{zhang2019bertscore}, \textit{COMET}\cite{rei2020comet}, \textit{METEOR}\cite{banerjee2005meteor}, and \textit{ROUGE}~\cite{Lin2004}. By leveraging these diverse measures, we aim to assess not only fluency and adequacy but also the preservation of technical content.

Specifically, this paper addresses the following research questions:

\begin{itemize}
\item \textbf{RQ1:} Which system most accurately identifies the source language?
\item \textbf{RQ2:} How do state-of-the-art machine translation (MT) models—such as Amazon Translate and DeepL—compare with large language models (LLMs) like ChatGPT, Claude, Gemini, LLaMA, and Mistral in preserving the technical precision of Visual Studio Code (VS Code) bug reports?

\end{itemize}

By answering these questions, we provide insights into the strengths and limitations of modern MT systems in software development contexts. Our findings aim to inform researchers and practitioners seeking to integrate automated translation into multilingual bug-triaging workflows effectively.

The remainder of this paper is structured as follows: Section~\ref{sec:related_work} contextualizes our work within the existing literature, Section~\ref{sec:data} describes our dataset, Section~\ref{sec:methodology} details our experimental methodology, Section~\ref{sec:results} presents results and analysis, and Section~\ref{sec:conclusion} offers concluding remarks and discussion.

\section{Related Work}
\label{sec:related_work}
Research on machine translation (MT) has advanced considerably over the past two decades, transitioning from rule-based and statistical paradigms to sophisticated neural approaches. Early neural models introduced by Bahdanau \emph{et al.}~\cite{bahdanau2014neural} demonstrated how attention mechanisms improve translation by aligning source and target words at each decoding step. Subsequent architectures, notably the Transformer~\cite{Vaswani2017}, showcased even more significant gains through multi-head attention and parallelizable layers. These innovations have substantially improved translation quality for general-domain text.

Despite the progress in neural MT, domain adaptation remains a critical challenge. Chu and Wang~\cite{Chu2018} provide a comprehensive survey of strategies for adapting neural models to specialized domains. They note that while generic training corpora enhance fluency and coverage, specialized domains—such as software engineering—require customized approaches to handle domain-specific lexicons and contexts effectively. Domain adaptation also affects translation evaluation, as metrics calibrated on general text may not accurately reflect correctness in specialized scenarios~\cite{Koehn2010}.

Within software engineering, bug reports constitute a unique category of text that often includes code snippets, API references, stack traces, and project-specific jargon. For instance, Arnaoudova \emph{et al.}~\cite{arnaoudova2016linguistic} highlights the linguistic complexity of software artifacts, emphasizing how ambiguities or errors in terminology can hinder comprehension and resolution of software issues.

Evaluating MT quality for these specialized texts necessitates both classic and modern metrics. BLEU~\cite{papineni2002bleu}, METEOR~\cite{banerjee2005meteor}, and ROUGE~\cite{Lin2004} have been widely adopted for general-purpose translation and summarization tasks. However, more recent metrics such as BERTScore~\cite{zhang2019bertscore} and COMET~\cite{rei2020comet} offer improved semantic sensitivity by leveraging transformer-based models for embedding and contextual analysis. These modern metrics may be particularly beneficial in software engineering contexts, where preserving the precise meaning of technical details is essential for successful bug resolution.

Finally, while industry-driven solutions such as AWS Translate, DeepL, and LLMs are commonly employed for multilingual documentation and user support, published comparative analyses specific to software bug reports remain limited. As such, a deeper examination of how these systems handle authentic bug reports is necessary and timely. Our work aims to fill this gap by systematically evaluating these MT solutions on bug reports from the Visual Studio Code repository, offering insights into their strengths and limitations in real-world software engineering.

\section{Dataset}
\label{sec:data}
\begin{figure}[htbp]
    \centering
    \includegraphics[width=0.48\textwidth]{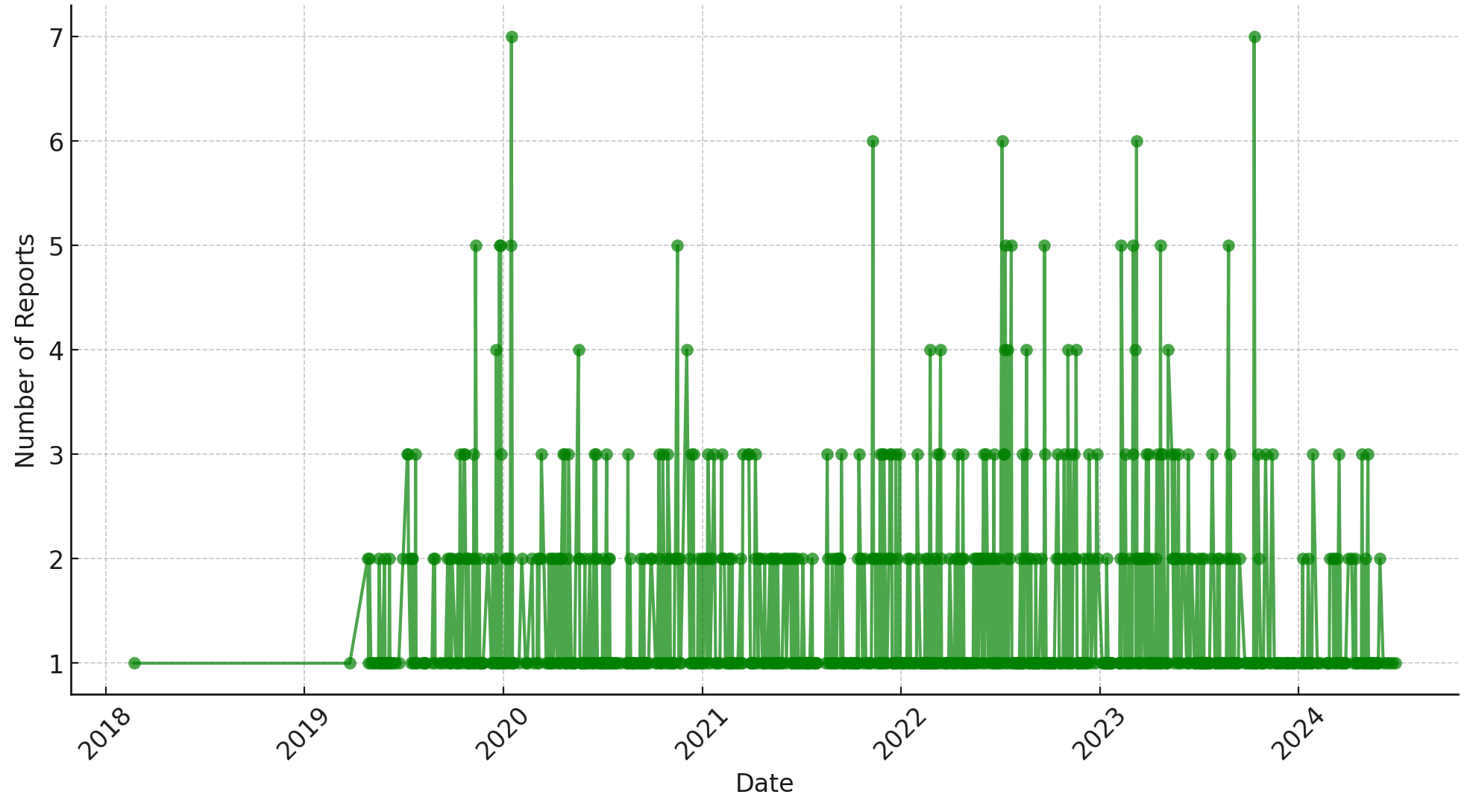}
    \caption{Bug reports over time.}
    \label{fig:br_ot}
\end{figure}

\begin{figure}[htbp]
    \centering
    \includegraphics[width=0.48\textwidth]{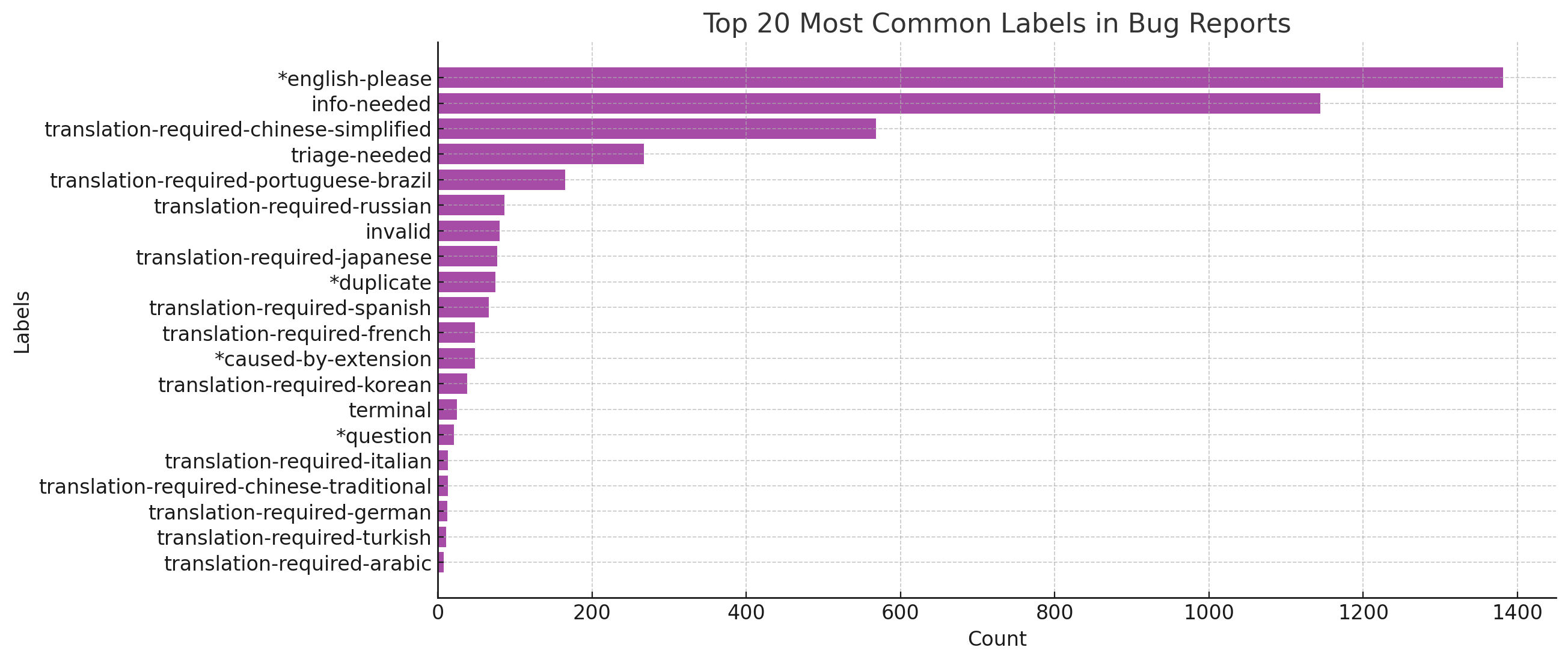}
    \caption{Top 20 most common labels in bug reports.}
    \label{fig:top20l}
\end{figure}

The dataset used in this study comprises bug reports extracted from the Visual Studio Code GitHub repository~\cite{VSCodeRepo}, specifically those labeled \textit{english-please}—a tag typically indicating submissions originally written in a language other than English, and thus serving as a reliable marker for multilingual content. Spanning five years (March 2019–June 2024), the dataset captures a wide variety of bug types, user environments, and technical contexts, as illustrated in \textbf{Figure~\ref{fig:br_ot}}. Although most \textit{english-please} issues were indeed non-English, some were already in English; these were retained to reflect real-world labeling inconsistencies and maintain dataset authenticity, as shown in \textbf{Figure~\ref{fig:top20l}}. The study focuses primarily on non-English reports, consistent with the tag’s intended purpose, to accurately represent the multilingual challenges in developer communication. The dataset’s linguistic diversity is summarized in \textbf{Table~\ref{tab:lang_distribution}}.

Each selected issue was paired with a reference English translation, verified by one or more bilingual contributors. This process yielded a final corpus of 1,300 multilingual bug reports covering a broad range of functional, UI, and performance-related issues. Given that bug reports vary widely in technical detail and domain-specific language, the dataset was curated to preserve this diversity and provide a robust testbed for evaluating the translation performance of all models considered.

\begin{table}[htbp]
\caption{Language distribution of bug reports.}
\label{tab:lang_distribution}
\centering
\renewcommand{\arraystretch}{1.2} 
\begin{tabular}{l r}
\hline
\textbf{Language} & \textbf{Number of Reports} \\
\hline
\multicolumn{2}{l}{\textbf{Most common languages}} \\
Chinese (Simplified) (zh-CN) & 637 \\
Portuguese (pt) & 151 \\
English (en) & 141 \\
Japanese (ja) & 76 \\
Russian (ru) & 75 \\
\hline
\multicolumn{2}{l}{\textbf{Less frequent languages (10--66)}} \\
\multicolumn{2}{p{6cm}}{Spanish (es), French (fr), Korean (ko), Traditional Chinese (zh-TW), German (de), Italian (it), Turkish (tr)} \\
\hline
\multicolumn{2}{l}{\textbf{Rare languages (\textless5)}} \\
\multicolumn{2}{p{6cm}}{Persian (fa), Ukrainian (uk), Polish (pl), Czech (cs), Arabic (ar), Indonesian (id), Serbian (sr)} \\
\hline
\end{tabular}
\end{table}

\section{Methodology}
\label{sec:methodology}

This section outlines our approach to evaluating the translation and language identification performance of seven state-of-the-art machine translation (MT) models on a curated dataset of 1,300 multilingual bug reports. We tested both commercial services (AWS Translate, DeepL, ChatGPT) and research-grade LLMs (Claude, Gemini, Mistral, LLaMA) under consistent conditions. Our evaluation combines translation quality metrics—BLEU, METEOR, ROUGE, BERTScore, and COMET—with classification measures such as accuracy, precision, recall, F1-score, and confusion matrix analysis. This methodology enables a comprehensive comparison of how effectively each model handles both technical translation and source language detection in a multilingual software development setting.


\subsection{Translation Systems}
\label{sec:translation-systems}

We evaluated seven state-of-the-art machine translation (MT) models, encompassing both commercial services and open-source large language models (LLMs):

\begin{enumerate}
    \item \textbf{AWS Translate:} A production-grade MT service, AWS Translate offers scalable, general-purpose translations. It performs reliably on standard text but may encounter challenges when translating domain-specific content such as developer jargon, inline code, or structured formatting~\cite{Chu2018}.

    \item \textbf{DeepL:} DeepL is widely recognized for producing fluent and coherent translations, particularly for European languages. It leverages a Transformer-based architecture optimized for natural language generation. However, it may inconsistently handle technical constructs such as method names, variable identifiers, and embedded code~\cite{arnaoudova2016linguistic}.

    \item \textbf{ChatGPT (GPT-4o):} Accessed via the OpenAI API, ChatGPT is a general-purpose large language model built on Transformer-based architectures~\cite{Vaswani2017}. It supports multilingual translation through prompt-based interaction and is capable of handling complex linguistic phenomena. However, performance may vary due to periodic updates to the underlying model.

    \item \textbf{Claude (Sonnet 3.7):} Developed by Anthropic, Claude is an LLM designed with a focus on safety and contextual understanding. It demonstrates strong multilingual capabilities and emphasizes the preservation of meaning, which is particularly beneficial for maintaining the intent of technical descriptions in bug reports.

    \item \textbf{Gemini (Pro 1.5):} Gemini (formerly Bard), developed by Google DeepMind, combines multilingual translation abilities with large-scale knowledge integration. It excels in precision during language identification and is capable of responding to natural-language prompts with detailed translations.

    \item \textbf{Mistral (Pixtral Large):} Mistral is an open-weight, decoder-only Transformer model engineered for high efficiency and multilingual performance. It demonstrates strong recall and F1-scores in language identification tasks, suggesting robust coverage across diverse linguistic patterns.

    \item \textbf{LLaMA (Maverick):} LLaMA (Large Language Model Meta AI), developed by Meta, is a family of open-source foundation models trained on diverse multilingual corpora. Although not explicitly optimized for translation tasks, it performs competitively in multilingual settings due to its extensive pretraining on varied textual data.
\end{enumerate}

We submitted each of the 1,300 bug reports to these three systems via their respective APIs. To maintain fairness, we preserved all formatting (code blocks, bullet points), and no system received additional context beyond the raw bug description.

\subsection{Evaluation Metrics}
\label{sec:eval_metrics}

Following best practices in MT research, we employed five automatic metrics to capture different facets of translation quality:

\begin{itemize}
    \item \textbf{BLEU}~\cite{papineni2002bleu}: A precision-oriented metric that measures $n$-gram overlaps between the system translation and the reference. Despite its popularity, BLEU can sometimes undervalue semantic coherence in specialized domains~\cite{Koehn2010}.

    \vspace{6pt}
    \noindent
    \textit{Formulation.}
    BLEU computes modified $n$-gram precision and multiplies it by a brevity penalty (BP) to discourage overly short hypotheses. Given a candidate translation $C$ and one or more references $R$:
    \begin{align}
        p_n &= \frac{\sum_{\text{n-gram}\in C} \min\bigl(\text{Count}_{C}(\text{n-gram}), \text{Count}_{R}(\text{n-gram})\bigr)}
                    {\sum_{\text{n-gram}\in C} \text{Count}_{C}(\text{n-gram})}, \\[6pt]
        \mathrm{BP} &= 
        \begin{cases}
            1, & \text{if } c > r,\\
            e^{1 - \tfrac{r}{c}}, & \text{if } c \le r,
        \end{cases}
    \end{align}
    where $c$ is the length of the candidate and $r$ is the effective reference length. The overall BLEU (often BLEU-4) is then:
    \begin{equation}
        \mathrm{BLEU} \;=\;
        \mathrm{BP} \;\times\; \exp\Bigl(\sum_{n=1}^{N} w_{n} \log p_n\Bigr),
    \end{equation}
    with $w_{n}$ typically set to $\tfrac{1}{N}$ and $N=4$.

    \item \textbf{METEOR}~\cite{banerjee2005meteor}: Improves upon BLEU by considering synonyms, stemming, and partial matches, thus often correlating better with human judgment.

    \vspace{6pt}
    \noindent
    \textit{Formulation.}
    METEOR first aligns unigrams from candidate and reference (accounting for stems, synonyms, and exact matches), then computes:
    \begin{align}
        P &= \frac{\text{\# of matched unigrams}}{\text{\# of unigrams in candidate}}, \\[6pt]
        R &= \frac{\text{\# of matched unigrams}}{\text{\# of unigrams in reference}}, \\[6pt]
        F_{\alpha} &= \frac{P R}{\alpha P + (1-\alpha)R}.
    \end{align}
    where $\alpha$ is often set to 0.9. A fragmentation penalty $\mathrm{Penalty}$, based on how the matched chunks are distributed, is applied:
    \begin{equation}
        \mathrm{Penalty} = \gamma \left(\frac{C}{M}\right)^\beta,
    \end{equation}
    where $C$ is the number of contiguous matched chunks, $M$ is the total number of matches, and $\gamma,\beta$ are hyperparameters (commonly $\gamma=0.5$, $\beta=3$). The final METEOR score is:
    \begin{equation}
        \mathrm{METEOR} = F_{\alpha}\, \bigl(1 - \mathrm{Penalty}\bigr).
    \end{equation}

    \item \textbf{ROUGE}~\cite{Lin2004}: Originally designed for summarization tasks, ROUGE assesses recall by tracking overlapping $n$-grams and sequences. Applied here, it highlights how comprehensively a translation covers the reference content.

    \vspace{6pt}
    \noindent
    \textit{Formulation (ROUGE-N).}
    The most common variant, ROUGE-N, measures the $n$-gram overlap. For a reference $R$ and candidate $C$:
    \newline
    
    \begin{equation}
        \begin{aligned}
            &\frac{
                \sum\limits_{\text{n-gram} \,\in\, R} 
                \min\bigl(\mathrm{Count}_{C}(\text{n-gram}), 
                          \mathrm{Count}_{R}(\text{n-gram})\bigr)
            }
            {
                \sum\limits_{\text{n-gram} \,\in\, R} 
                \mathrm{Count}_{R}(\text{n-gram})
            }
        \end{aligned}
    \end{equation}

    This yields a recall-oriented view (though precision or $F$-measure versions also exist).

    \item \textbf{BERT Score}~\cite{zhang2019bertscore}: Leverages contextual embeddings from Transformer-based models to measure semantic alignment. Particularly useful for identifying subtle meaning shifts in technical text.

    \vspace{6pt}
    \noindent
    \textit{Formulation.}
    Let $\mathbf{c}_j$ be the embedding of the $j$-th token in candidate $C$ and $\mathbf{r}_i$ the embedding of the $i$-th token in reference $R$. Using cosine similarity,
    \[
      s(\mathbf{c}_j, \mathbf{r}_i) \;=\; 
      \frac{\mathbf{c}_j \cdot \mathbf{r}_i}{\|\mathbf{c}_j\|\;\|\mathbf{r}_i\|}.
    \]
    Then BERT Score precision and recall are:
    \begin{align}
      \mathrm{Precision} &= 
        \frac{1}{|C|} \sum_{c_j \in C} \max_{r_i \in R}\,s(\mathbf{c}_j,\mathbf{r}_i), \\[4pt]
      \mathrm{Recall} &=
        \frac{1}{|R|} \sum_{r_i \in R} \max_{c_j \in C}\,s(\mathbf{r}_i,\mathbf{c}_j),
    \end{align}
    and the $F_1$ score is
    \[
      \mathrm{F1} = \frac{2\,(\mathrm{Precision} \times \mathrm{Recall})}
                         {\mathrm{Precision} + \mathrm{Recall}}.
    \]
    Reported BERT Score is typically this F1 value.

    \item \textbf{COMET}~\cite{rei2020comet}: A newer neural-based framework that ranks translations through pretrained language models, offering a more holistic approach to semantic and syntactic quality evaluation.

    \vspace{6pt}
    \noindent
    \textit{Formulation (Conceptual).}
    COMET uses a pretrained encoder (e.g., XLM-R) to obtain embeddings for the source $x$, hypothesis (candidate) $y$, and reference $r$:
    \[
      \mathbf{h}_x = \mathrm{Enc}(x), \quad
      \mathbf{h}_y = \mathrm{Enc}(y), \quad
      \mathbf{h}_r = \mathrm{Enc}(r).
    \]
    These embeddings are fed into a learned regressor $\phi_W$:
    \[
      \hat{Q}(x,y,r) \;=\; \phi_{W}\!\bigl(\mathbf{h}_x, \mathbf{h}_y, \mathbf{h}_r\bigr),
    \]
    producing a quality estimate. The final COMET score is:
    \begin{equation}
      \mathrm{COMET}(x,y,r) \;=\; \hat{Q}(x,y,r).
    \end{equation}
    Since $\phi_W$ is trained on human judgments, COMET effectively learns to predict translation quality across diverse contexts.

    \item \textbf{Accuracy}\cite{manning2008introduction}: A fundamental metric used to evaluate classification models, measuring the proportion of correctly predicted instances over the total number of instances.
    
    \vspace{6pt}
    \noindent
    \textit{Formulation.}  
    Given a classification task with predictions $\hat{y}$ and ground-truth labels $y$, accuracy is defined in terms of true positives (TP), true negatives (TN), false positives (FP), and false negatives (FN):
    \[
      \mathrm{Accuracy} = \frac{\text{TP} + \text{TN}}{\text{TP} + \text{TN} + \text{FP} + \text{FN}}.
    \]
    This metric provides an overall measure of correctness but may be insufficient for imbalanced datasets. 
    
    For multi-class classification with $N$ samples and $K$ classes, accuracy extends to:
    \[
      \mathrm{Accuracy} = \frac{1}{N} \sum_{i=1}^{N} \mathbb{1}(\hat{y}_i = y_i),
    \]
    where $\mathbb{1}(\cdot)$ is the indicator function that returns 1 if the prediction matches the true label, otherwise 0.

    \item \textbf{Precision, Recall, and F1-score}\cite{manning2008introduction}: These metrics provide deeper insight into classification performance, especially for imbalanced datasets.

    \begin{itemize}
        \item \textbf{Precision (per language)}: Measures how many predicted labels for a given language are actually correct.
        \[
        \mathrm{Precision} = \frac{\text{TP}}{\text{TP} + \text{FP}}.
        \]
        
        \item \textbf{Recall (per language)}: Measures how many actual labels for a given language were correctly predicted.
        \[
        \mathrm{Recall} = \frac{\text{TP}}{\text{TP} + \text{FN}}.
        \]

        \item \textbf{F1-score}: The harmonic mean of precision and recall, balancing both metrics.
        \[
        \mathrm{F1} = 2 \times \frac{\mathrm{Precision} \times \mathrm{Recall}}{\mathrm{Precision} + \mathrm{Recall}}.
        \]

    \end{itemize}

    \item \textbf{Confusion Matrix}: A structured table that helps visualize how often each language is correctly classified or misclassified as another. The confusion matrix for $K$ classes is defined as:

    \[
    C_{i,j} = \text{Number of instances of class } i \text{ classified as class } j.
    \]

    The diagonal elements represent correct predictions, while off-diagonal elements indicate misclassifications. This visualization is especially useful for understanding error patterns across different languages.

\end{itemize}

\section{Results}
\label{sec:results}
\begin{figure*}[htbp]
    \centering
    \caption{Violin plots of different MT evaluation metrics across AWS, GPT, and DeepL translation tools.}
    \begin{subfigure}[b]{0.32\textwidth}
        \includegraphics[width=\textwidth]{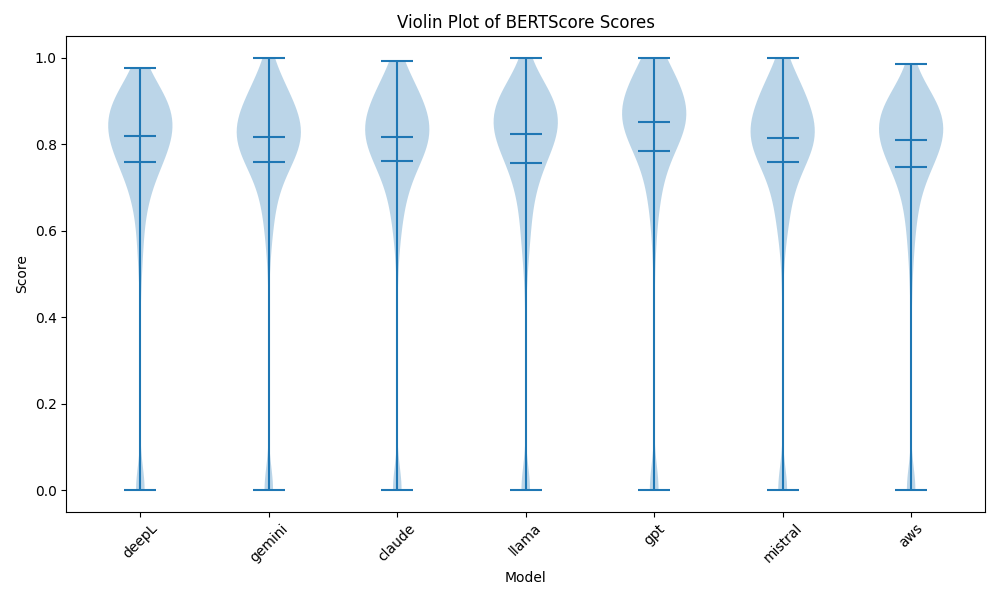}
        \caption{BERT Scores}
        \label{fig:bert}
    \end{subfigure}
    \begin{subfigure}[b]{0.32\textwidth}
        \includegraphics[width=\textwidth]{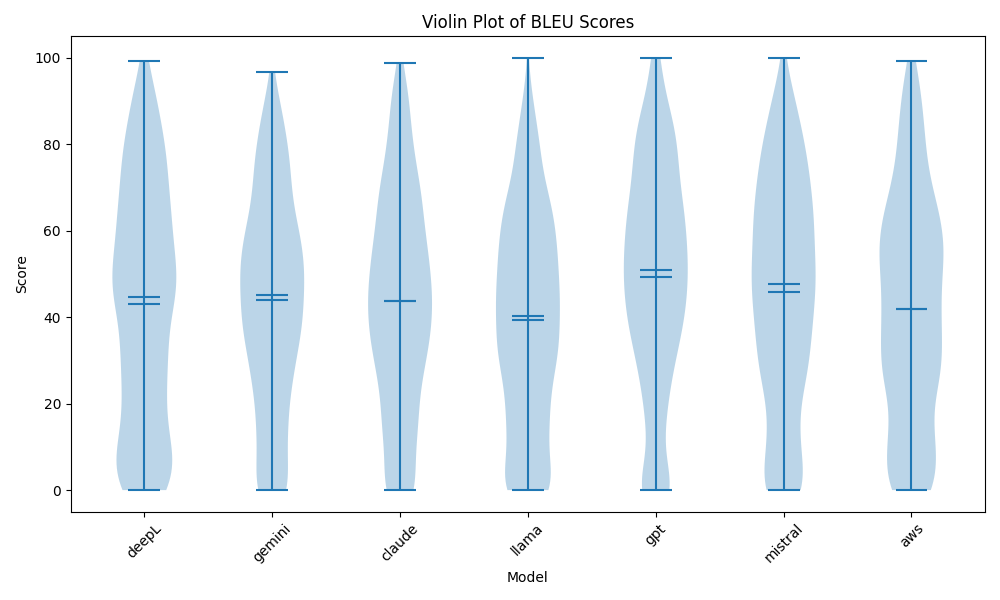}
        \caption{BLEU Scores}
        \label{fig:bleu}
    \end{subfigure}
    \begin{subfigure}[b]{0.32\textwidth}
        \includegraphics[width=\textwidth]{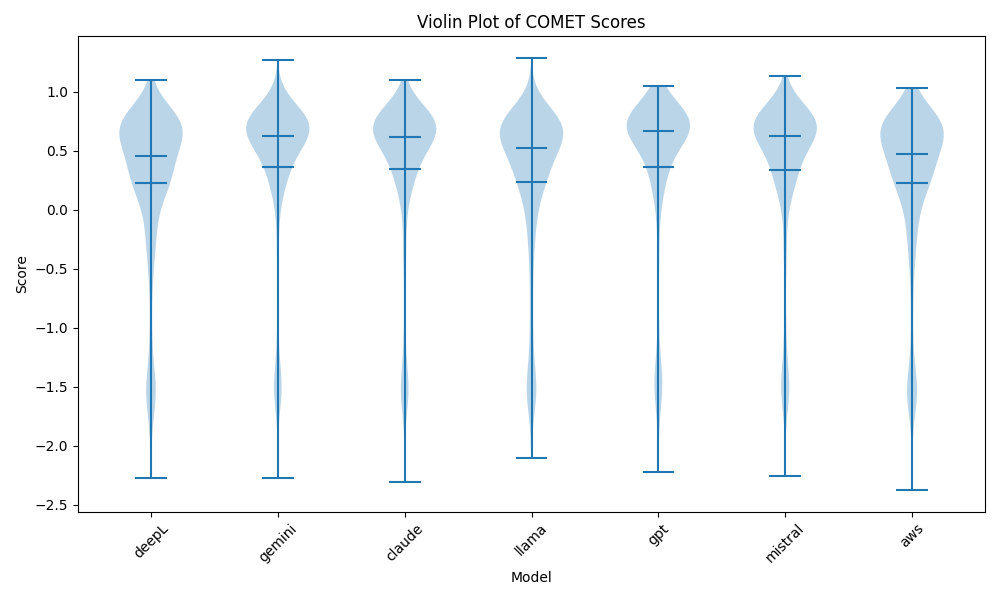}
        \caption{COMET Scores}
        \label{fig:comet}
    \end{subfigure}
    \begin{subfigure}[b]{0.32\textwidth}
        \includegraphics[width=\textwidth]{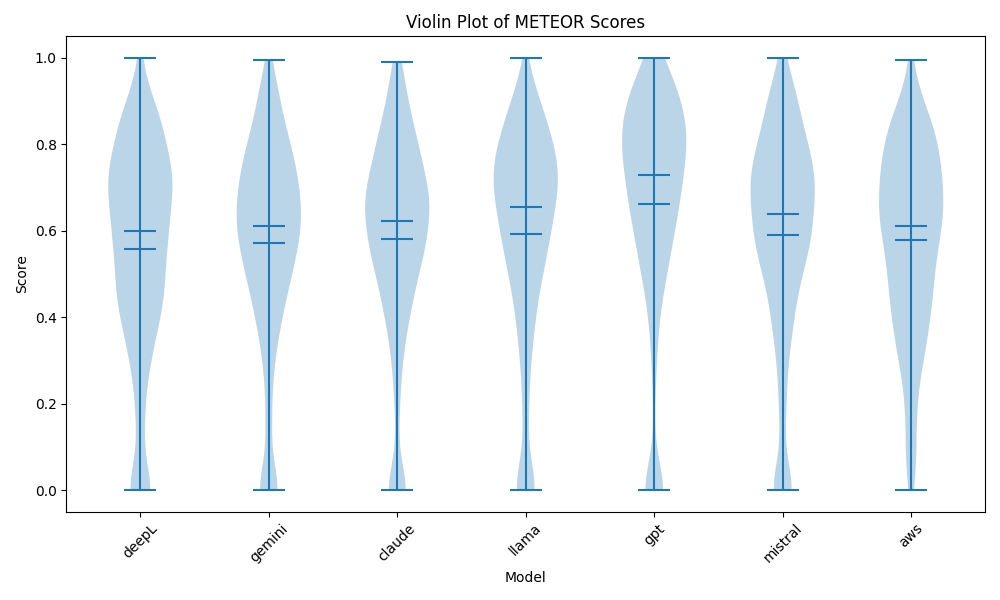}
        \caption{METEOR Scores}
        \label{fig:meteor}
    \end{subfigure}
    \begin{subfigure}[b]{0.32\textwidth}
        \includegraphics[width=\textwidth]{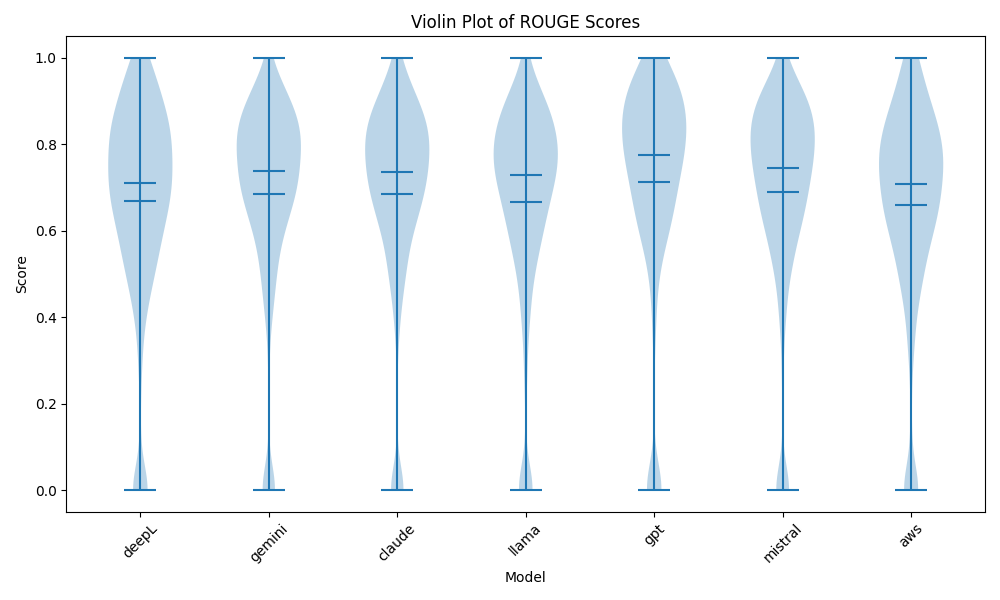}
        \caption{ROUGE-L Scores}
        \label{fig:rouge}
    \end{subfigure}

    \label{fig:violin_plots}
\end{figure*}

This section presents the findings of our evaluation of ChatGPT, AWS Translate, and DeepL on the curated bug report dataset. Figure~\ref{fig:violin_plots} presents the distribution of translation model scores using violin plots.

\subsection{Automatic Metrics}
\label{subsec:auto_metrics}

Table~\ref{tab:auto_metrics} summarizes the results of five automatic evaluation metrics—BLEU, METEOR, ROUGE, BERTScore, and COMET—computed over 1,300 multilingual bug reports. The highest scores for each metric are highlighted in bold, and models are listed in descending order of overall performance.

\begin{table}[htbp]
    \centering
    \caption{Automatic metric scores of translation systems.}
    \label{tab:auto_metrics}
    \renewcommand{\arraystretch}{1.2} 
    \begin{tabular}{lccccc}
        \toprule
        \textbf{Model} & \textbf{BERTScore} & \textbf{BLEU} & \textbf{COMET} & \textbf{METEOR} & \textbf{ROUGE} \\
        \midrule
        GPT     & \textbf{0.7838} & \textbf{49.32} & \textbf{0.3636} & \textbf{0.6613} & \textbf{0.7128} \\
        Mistral & 0.7597 & 45.85 & 0.3325 & 0.5895 & 0.6903 \\
        Claude  & 0.7626 & 43.68 & 0.3417 & 0.5807 & 0.6861 \\
        Gemini  & 0.7595 & 43.87 & 0.3604 & 0.5721 & 0.6854 \\
        DeepL   & 0.7589 & 43.06 & 0.2289 & 0.5570 & 0.6680 \\
        Llama   & 0.7576 & 39.42 & 0.2313 & 0.5923 & 0.6661 \\
        AWS     & 0.7487 & 41.90 & 0.2288 & 0.5792 & 0.6599 \\
        \bottomrule
    \end{tabular}
    \label{tab:evaluation_scores}
    \end{table}

\noindent\textbf{BLEU}: ChatGPT (gpt-4o) achieved the highest BLEU score (49.32), followed by Mistral (45.85) and Gemini (43.87), indicating that ChatGPT had the greatest n-gram overlap with the reference translations. This suggests strong surface-level alignment, though BLEU may still fall short in capturing deeper semantic equivalence, particularly in technical text.

\noindent\textbf{METEOR}: ChatGPT also led in METEOR (0.6613), outperforming LLaMA (0.5923) and Mistral (0.5895), suggesting superior handling of synonyms, morphological variants, and paraphrases. This indicates that ChatGPT's outputs were more aligned with the reference in terms of linguistic diversity and fluency.

\noindent\textbf{ROUGE}: ChatGPT recorded the highest ROUGE-L F1 score (0.7128), followed by Mistral (0.6903) and Claude (0.6861). Given the technical nature of bug reports, where exact phrasing often recurs, this result reflects ChatGPT’s strength in preserving key terminologies and phrasings.

\noindent\textbf{BERTScore}: ChatGPT obtained the highest BERTScore (0.7838), indicating the strongest semantic alignment with the reference translations. Claude (0.7626) and Gemini (0.7595) followed closely, further supporting ChatGPT’s contextual accuracy in domain-specific content.

\noindent\textbf{COMET}: ChatGPT again topped the scores with 0.3636, followed by Gemini (0.3604) and Claude (0.3417). As COMET leverages pretrained models for semantic assessment, these scores reinforce ChatGPT’s ability to preserve meaning in complex, technical inputs.

\subsection{Source Language Identification}
\label{subsec:lang_id}

Table~\ref{tab:translation_comparison} compares the language identification performance of seven translation models. AWS Translate achieved the highest accuracy (0.4717), indicating its strength in correctly identifying the source language. However, Gemini exhibited the highest precision (0.7414), meaning it was most likely to be correct when it did make a prediction. Claude led in recall (0.7581), capturing the largest proportion of actual positive instances, and also achieved the highest F1-score (0.7182), reflecting a strong balance between precision and recall. Among all models, GPT, Mistral, LLaMA, and Gemini demonstrated comparably high precision and recall, highlighting their consistency across predictions. DeepL maintained a balanced performance across all metrics, with an F1-score of 0.6716. The confusion matrix is visualized in \textbf{Figure~\ref{fig:cm_plots}}.

\begin{table}[ht]
    \centering
    \caption{Language identification scores.}
    \label{tab:translation_comparison}
    \renewcommand{\arraystretch}{1.2} 
    \begin{tabular}{lcccc}
        \toprule
        \textbf{Model} & \textbf{Accuracy} & \textbf{Precision} & \textbf{Recall} & \textbf{F1-score} \\
        \midrule
        AWS     & \textbf{0.4717} & 0.6788 & 0.7488 & \textbf{0.7027} \\
        DeepL   & 0.4586 & 0.6598 & 0.6954 & 0.6716 \\
        Claude  & 0.4079 & 0.7129 & \textbf{0.7581} & \textbf{0.7182} \\
        GPT     & 0.4070 & 0.7130 & 0.7190 & 0.6788 \\
        Mistral & 0.4062 & 0.7120 & 0.7574 & \textbf{0.7142} \\
        LLaMA   & 0.4029 & 0.7098 & 0.7557 & 0.7120 \\
        Gemini  & 0.4005 & \textbf{0.7414} & 0.7639 & 0.7123 \\
        \bottomrule
    \end{tabular}
    \label{tab:classification_metrics}
\end{table}

\begin{figure*}[t]
    \centering
    \begin{subfigure}[b]{0.32\textwidth}
        \includegraphics[width=\textwidth]{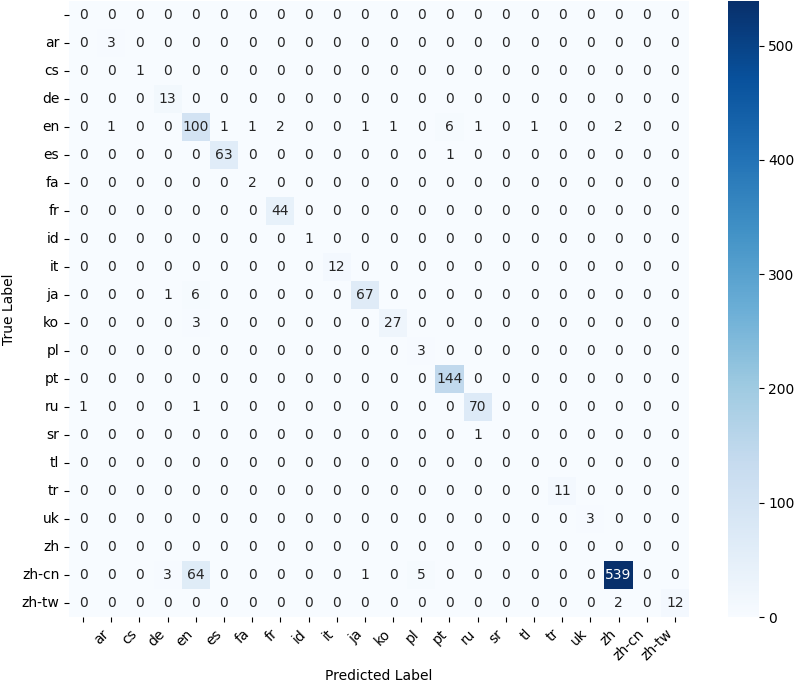}
        \caption{AWS Translate}
        \label{fig:aws_cm}
    \end{subfigure}
    \hfill
    \begin{subfigure}[b]{0.32\textwidth}
        \includegraphics[width=\textwidth]{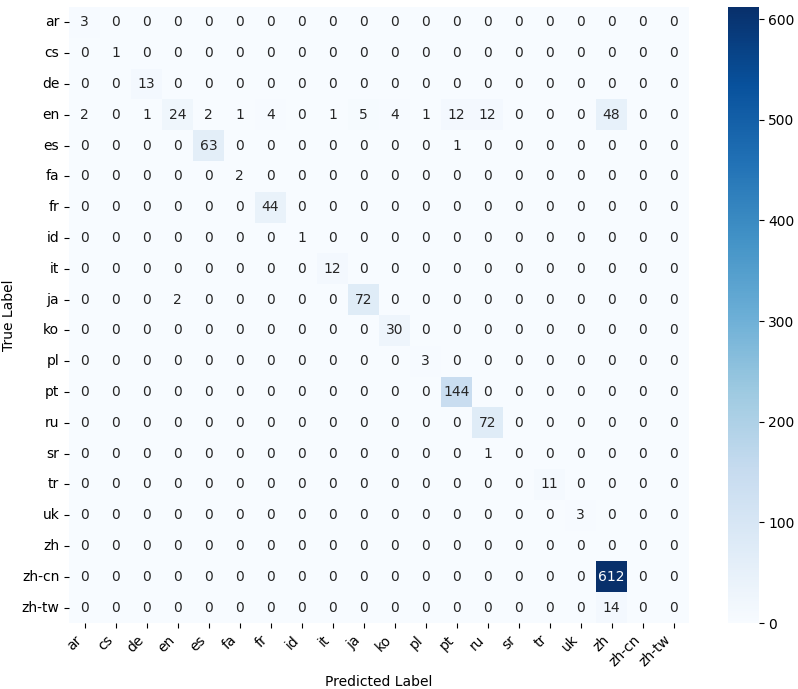}
        \caption{Claude}
        \label{fig:claude_cm}
    \end{subfigure}
    \hfill
    \begin{subfigure}[b]{0.32\textwidth}
        \includegraphics[width=\textwidth]{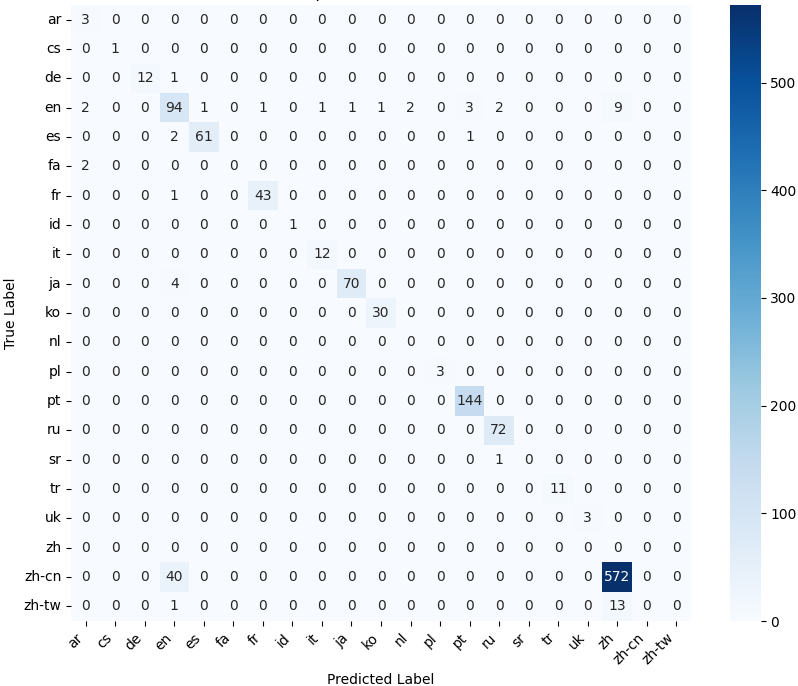}
        \caption{DeepL}
        \label{fig:deepl_cm}
    \end{subfigure}
    
    \vspace{0.2cm} 
    \begin{subfigure}[b]{0.32\textwidth}
        \includegraphics[width=\textwidth]{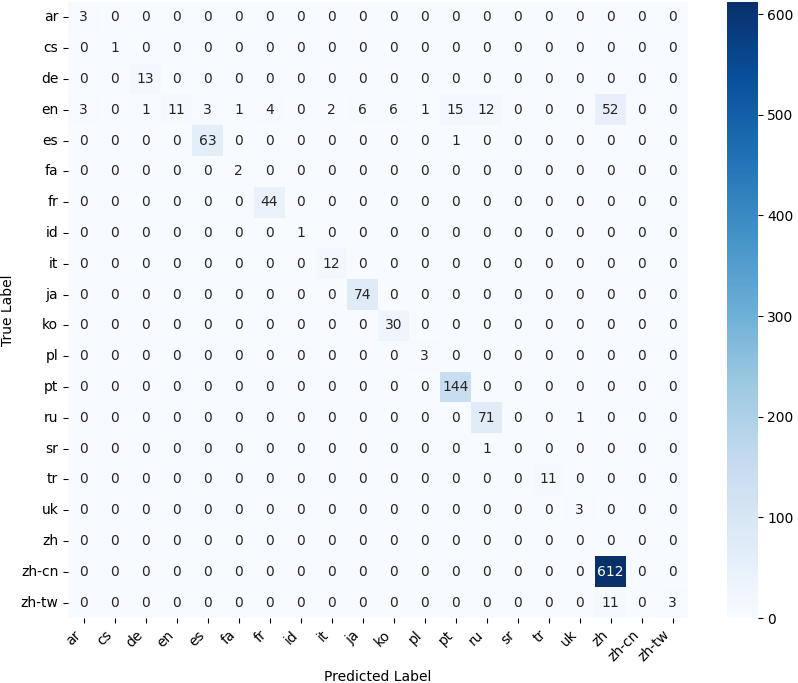}
        \caption{Gemini}
        \label{fig:gemini_cm}
    \end{subfigure}
    \hfill
    \begin{subfigure}[b]{0.32\textwidth}
        \includegraphics[width=\textwidth]{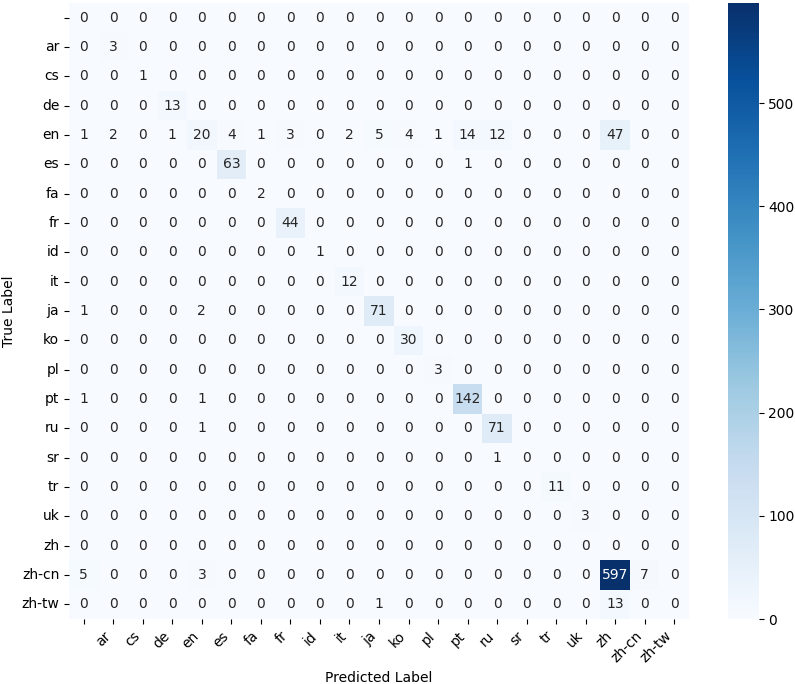}
        \caption{ChatGPT}
        \label{fig:gpt_cm}
    \end{subfigure}
    \hfill
    \begin{subfigure}[b]{0.32\textwidth}
        \includegraphics[width=\textwidth]{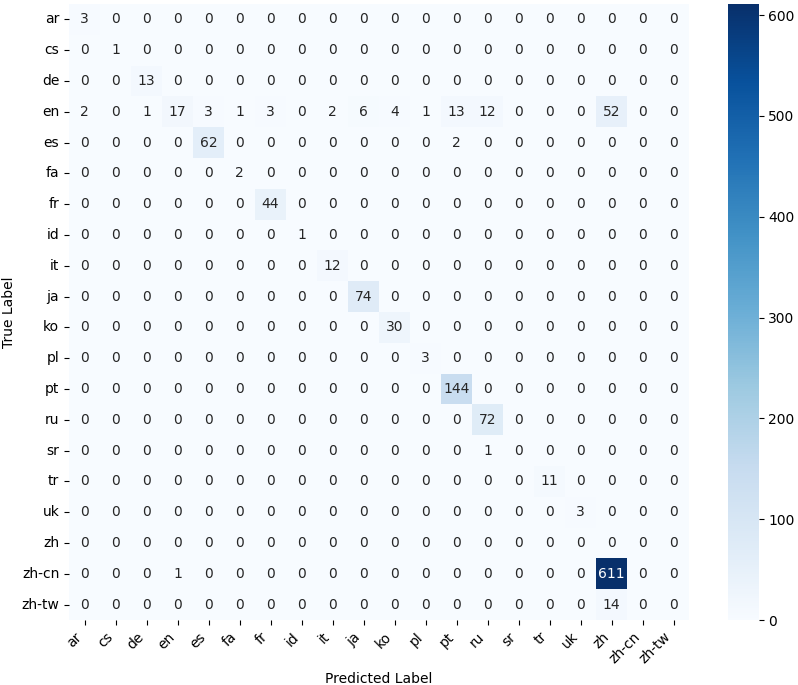}
        \caption{Llama}
        \label{fig:llama_cm}
    \end{subfigure}
    
    \vspace{0.2cm} 
    \centering
    \begin{subfigure}[b]{0.32\textwidth}
        \includegraphics[width=\textwidth]{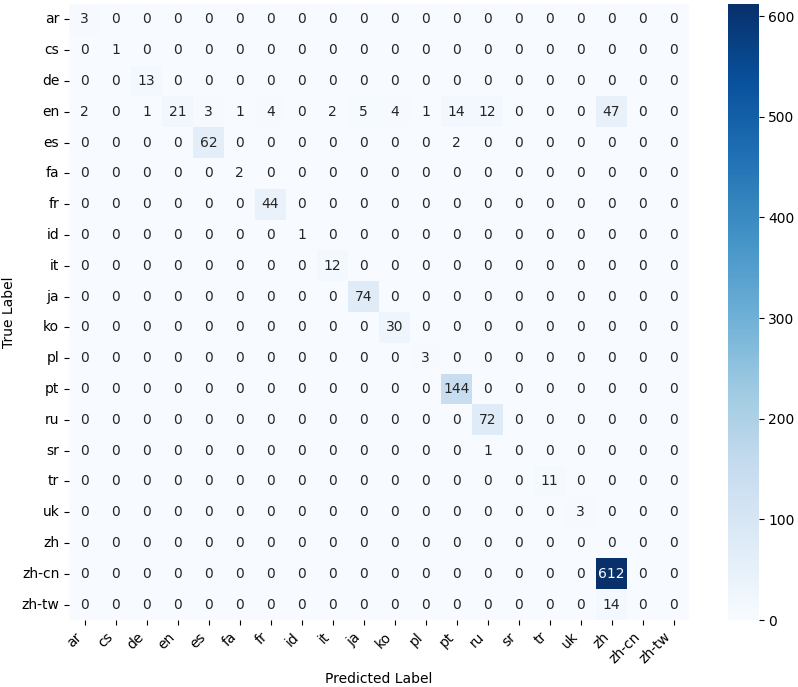}
        \caption{Mistral}
        \label{fig:mistral_cm}
    \end{subfigure}
    
    \caption{Confusion matrices for language identification across seven machine translation models. Darker diagonal entries indicate correct classifications.}
    \label{fig:confusion_matrices}
\end{figure*}

\subsection{Summary of Findings}
ChatGPT (gpt-4o) outperformed all other models across all five automatic translation metrics—BLEU, METEOR, ROUGE, BERTScore, and COMET—demonstrating strong surface-level accuracy and deep semantic alignment in multilingual bug report translations. Mistral and Gemini also performed competitively, particularly in BLEU and COMET, while Claude showed strong performance in BERTScore and ROUGE. DeepL and AWS Translate lagged behind in translation quality, indicating reduced semantic fidelity and n-gram correspondence.

For language identification, no single model dominated across all metrics. AWS Translate achieved the highest accuracy (0.4717), making it more reliable for correct classifications overall. However, Claude attained the highest F1-score (0.7182) and recall (0.7581), suggesting its strength in capturing relevant instances. Gemini led in precision (0.7414), favoring correctness in its predictions. The best model depends on the specific performance criterion prioritized—accuracy, precision, or recall.

\section{Discussion and Conclusion}
\label{sec:conclusion}

Our evaluation of leading machine translation (MT) systems—ChatGPT, Claude, Gemini, Mistral, DeepL, AWS Translate, and LLaMA—on multilingual bug reports from the Visual Studio Code repository reveals both promise and limitations in domain-specific translation.

ChatGPT achieved the highest overall performance across all automatic metrics (BLEU, METEOR, ROUGE, BERTScore, and COMET), demonstrating its capability to retain both lexical fidelity and semantic integrity in technical contexts. Claude, Gemini, and Mistral also delivered competitive results, affirming that large language models (LLMs) can generalize well to structured, domain-specific inputs when properly tuned. In contrast, DeepL and AWS Translate underperformed on most translation quality metrics, possibly due to insufficient domain adaptation. Nevertheless, AWS Translate excelled in source language identification (accuracy: 0.4717, F1-score: 0.7027), while Claude and Mistral achieved higher F1-scores (0.7182 and 0.7142, respectively), suggesting trade-offs between precision and consistency.

Discrepancies among traditional metrics (BLEU, METEOR, ROUGE) and embedding-based metrics (BERTScore, COMET) emphasize the limitations of $n$-gram approaches in capturing semantic nuances. ChatGPT’s consistent top-tier performance across all five metrics challenges earlier assumptions about LLM shortcomings in technical translation and underscores the value of context-aware evaluation.

It is important to note the evolving nature of cloud-based MT systems. Regular updates to models such as ChatGPT and AWS Translate may lead to temporal variability in translation performance. Consequently, our results represent a time-bound snapshot rather than a static benchmark. Longitudinal studies are needed to track the progression and stability of these models over time.

While our dataset offers a realistic benchmark through bug reports from a popular open-source project, it remains limited to a single codebase. Expanding to include diverse projects and multilingual annotations would enhance generalizability. Additionally, reliance on a small group of domain-knowledgeable annotators introduces the potential for subjective bias.

In summary, modern MT systems—particularly LLM-based models—demonstrate strong potential for translating technical content like bug reports. Yet, challenges remain in preserving domain-specific terminology and structural fidelity. Future work should focus on improving neural architectures, enhancing domain adaptation techniques, and integrating specialized linguistic resources to support software engineering tasks.

\bibliographystyle{IEEEtran}
\bibliography{ref}

\begin{thebibliography}{10}
\providecommand{\url}[1]{#1}
\csname url@samestyle\endcsname
\providecommand{\newblock}{\relax}
\providecommand{\bibinfo}[2]{#2}
\providecommand{\BIBentrySTDinterwordspacing}{\spaceskip=0pt\relax}
\providecommand{\BIBentryALTinterwordstretchfactor}{4}
\providecommand{\BIBentryALTinterwordspacing}{\spaceskip=\fontdimen2\font plus
\BIBentryALTinterwordstretchfactor\fontdimen3\font minus \fontdimen4\font\relax}
\providecommand{\BIBforeignlanguage}[2]{{%
\expandafter\ifx\csname l@#1\endcsname\relax
\typeout{** WARNING: IEEEtran.bst: No hyphenation pattern has been}%
\typeout{** loaded for the language `#1'. Using the pattern for}%
\typeout{** the default language instead.}%
\else
\language=\csname l@#1\endcsname
\fi
#2}}
\providecommand{\BIBdecl}{\relax}
\BIBdecl

\bibitem{chu2018survey}
C.~Chu and R.~Wang, ``A survey of domain adaptation for neural machine translation,'' \emph{arXiv preprint arXiv:1806.00258}, 2018.

\bibitem{bahdanau2014neural}
D.~Bahdanau, ``Neural machine translation by jointly learning to align and translate,'' \emph{arXiv preprint arXiv:1409.0473}, 2014.

\bibitem{Vaswani2017}
A.~Vaswani, N.~Shazeer, N.~Parmar, J.~Uszkoreit, L.~Jones, A.~N. Gomez, {\L}.~Kaiser, and I.~Polosukhin, ``Attention is all you need,'' in \emph{Advances in Neural Information Processing Systems (NIPS)}, 2017, pp. 5998--6008.

\bibitem{VSCodeRepo}
Microsoft, ``Visual studio code github repository,'' \url{https://github.com/microsoft/vscode}, 2023.

\bibitem{arnaoudova2016linguistic}
V.~Arnaoudova, M.~Di~Penta, and G.~Antoniol, ``Linguistic antipatterns: What they are and how developers perceive them,'' \emph{Empirical Software Engineering}, vol.~21, pp. 104--158, 2016.

\bibitem{papineni2002bleu}
K.~Papineni, S.~Roukos, T.~Ward, and W.-J. Zhu, ``Bleu: a method for automatic evaluation of machine translation,'' in \emph{Proceedings of the 40th annual meeting of the Association for Computational Linguistics}, 2002, pp. 311--318.

\bibitem{zhang2019bertscore}
T.~Zhang, V.~Kishore, F.~Wu, K.~Q. Weinberger, and Y.~Artzi, ``Bertscore: Evaluating text generation with bert,'' \emph{arXiv preprint arXiv:1904.09675}, 2019.

\bibitem{rei2020comet}
R.~Rei, C.~Stewart, A.~C. Farinha, and A.~Lavie, ``Comet: A neural framework for mt evaluation,'' \emph{arXiv preprint arXiv:2009.09025}, 2020.

\bibitem{banerjee2005meteor}
S.~Banerjee and A.~Lavie, ``Meteor: An automatic metric for mt evaluation with improved correlation with human judgments,'' in \emph{Proceedings of the acl workshop on intrinsic and extrinsic evaluation measures for machine translation and/or summarization}, 2005, pp. 65--72.

\bibitem{Lin2004}
C.-Y. Lin, ``{ROUGE}: A package for automatic evaluation of summaries,'' in \emph{Text Summarization Branches Out}.\hskip 1em plus 0.5em minus 0.4em\relax Barcelona, Spain: Association for Computational Linguistics, Jul. 2004, pp. 74--81.

\bibitem{Chu2018}
C.~Chu and R.~Wang, ``A survey of domain adaptation for neural machine translation,'' in \emph{Proceedings of the 27th International Conference on Computational Linguistics (COLING)}, 2018, pp. 1304--1319.

\bibitem{Koehn2010}
P.~Koehn, \emph{Statistical Machine Translation}.\hskip 1em plus 0.5em minus 0.4em\relax Cambridge University Press, 2010.

\bibitem{manning2008introduction}
\BIBentryALTinterwordspacing
C.~D. Manning, P.~Raghavan, and H.~Sch{\"u}tze, \emph{Introduction to Information Retrieval}.\hskip 1em plus 0.5em minus 0.4em\relax New York, NY, USA: Cambridge University Press, 2008. [Online]. Available: \url{https://nlp.stanford.edu/IR-book/}
\BIBentrySTDinterwordspacing

\end{thebibliography}
\vspace{12pt}

\end{document}